\documentclass[english]{cccconf}
\usepackage[comma,numbers,square,sort&compress]{natbib}
\usepackage{epstopdf}
\usepackage{bm}
\usepackage{amsmath}
\usepackage{booktabs}
\usepackage{subfigure}
\usepackage{multirow}
\usepackage{array}
\begin{document}

\title{Analytic Task Scheduler: Recursive Least Squares Based Method for Continual Learning in Embodied Foundation Models}

\author{Lipei Xie, Yingxin Li, Huiping Zhuang}



\affiliation{Shien-Ming Wu School of Intelligent Engineering, South China University of Technology, Guangzhou 510641, P.~R.~China\email{hpzhuang@scut.edu.cn}}

\maketitle

\begin{abstract}
Embodied foundation models are crucial for Artificial Intelligence (AI) interacting with the physical world by integrating multi-modal inputs, such as proprioception, vision and language, to understand human intentions and generate actions to control robots. While these models demonstrate strong generalization and few-shot learning capabilities, they face significant challenges in continually acquiring new skills without forgetting previously learned skills, a problem known as catastrophic forgetting. To address this issue, we propose the Analytic Task Scheduler (ATS), a novel framework for continual learning in embodied foundation models. ATS consists of a task-specific model library, where each model is fine-tuned independently on a single task, and an analytic scheduler trained using recursive least squares (RLS) to learn the mapping between language instructions and task-specific models. This architecture enables accurate task recognition and dynamic model selection while fundamentally avoiding parameter interference across tasks. The scheduler updates its parameters incrementally using only statistics (autocorrelation and cross-correlation matrices), enabling forgetting-resistant learning without the need to revisit historical data. We validate ATS on a real-world robot platform (RM65B), demonstrating superior resistance to forgetting and strong adaptability to task variations. The results highlight ATS as an effective, scalable, and deployable solution for continual learning in embodied foundation models operating in complex, dynamic environments. Our code will be available at https://github.com/MIAA-Embodied-AI/AnalyticTaskScheduler
\end{abstract}

\keywords {Embodied Foundation Models, Continual Learning, Recursive Least Squares Method, Analytic Task Scheduler} 


\section{Introduction}

Embodied foundation models\cite{1.1,1.2,1.3} integrate multi-modal inputs (proprioception, RGB-D images, language) to understand human intentions and environments, generating low-level actions to control robots. Recent years have witnessed significant advances in the development of embodied foundation models. These models adopt the training paradigm of “pre-training and fine-tuning”: pre-training on large, diverse datasets for general knowledge, and fine-tuning with task-specific data to enhance the adaptability of the model to specific application scenarios, ensuring smooth and efficient performance on downstream tasks. This training paradigm provides strong generalization, few-shot and zero-shot learning capabilities, facilitating the development of general-purpose embodied foundation models capable of performing complex tasks across various scenes.

However, the environments faced by Embodied Artificial Intelligence (Embodied AI) are typically open and dynamic, with task demands continuously evolving and new scenes and objectives potentially emerging at any time. As a result, the embodied foundation model is often unable to learn all the tasks through a single training phase. To maintain robust performance in such non-stationary environments, the embodied foundation model urgently needs to be equipped with continual learning capabilities. Continual learning, also known as incremental learning and lifelong learning, aims to enables the model to acquire new knowledge incrementally while retaining previously learned knowledge\cite{1.4}. Currently, a persistent challenge in this area is catastrophic forgetting\cite{1.5}, in which the performance of the model on previous tasks drops significantly after learning a new one. For embodied foundation models, traditional continual learning methods face several challenges, including increased computational and time overhead, potential violations of data privacy, and conflicts in parameter importance matrices.

To enhance the adaptability of embodied foundation models to new tasks, a widely adopted approach is to fine-tune the pre-trained model on small and high-quality datasets. This strategy enables rapid adaptation to downstream tasks without costly large-scale retraining, improving task-specific performance. Fine-tuning can be performed in two modes: single-task and multi-task. Single-task fine-tuning optimizes the model for a specific task, often achieving superior performance on that task. In contrast, multi-task fine-tuning jointly optimizes model parameters across tasks, promoting generalization and transfer but risking performance degradation due to cross-task interference. While both fine-tuning strategies offer practical benefits at specific stages, they fall short of supporting continual learning of embodied foundation models in real-world scenarios. Although single-task fine-tuning is task-specific, the model requires to be trained for each new task, lacking a unified mechanism for knowledge integration and lifelong adaptation, leading to fragmented learning and extensible long-term capabilities. While multi-task fine-tuning builds a unified model across tasks, incrementally fine-tuning an already fine-tuned multi-task model with new task data often causes catastrophic forgetting and degrades previous task performance, with growing task interference further destabilizing the model.

To address the limitations discussed above, we propose the Analytic Task Scheduler (ATS), a continual learning framework designed to learn new task in a replay-free way without destroying the original generic knowledge of the model, while effectively overcoming catastrophic forgetting. Specifically, to avoid interference between tasks and maintain pre-trained general knowledge, ATS adopts a single-task fine-tuning strategy to train an independent model for each task and gradually form a task-specific model library. This design ensures high task-specific performance while avoiding cross-task interference. In addition, to support continual learning for the model, ATS introduces RLS in the language input channel of the embodied foundation model to train an analytic scheduler, which enables it to incrementally learn to map language instructions to task-specific models.  The RLS algorithm plays a critical role in preserving prior knowledge without revisiting previous data, which is essential for ATS’s resistance to catastrophic forgetting. As a result, by continuously adapting the scheduler through language input, ATS progressively improves its capacity for task recognition and model selection. Coupled with the incrementally constructed model library, ATS forms a unified and scalable continual learning framework for embodied foundation models. Given a language instruction, ATS can accurately identify the associated task and dynamically routes the appropriate model to control the robot. The main contributions of this work are as follows:
\begin{itemize}
  \item We propose ATS, a continual learning framework for embodied foundation models that enables the dynamic learning of new task, significantly improving adaptability in evolving environments.
  \item Our approach fine-tunes a task-specific model for each task, preserving the general knowledge of the original foundation model while minimizing task interference and ensuring targeted optimization for each task.
  \item We introduce a RLS-based scheduler that incrementally learns new tasks, enabling continual learning without catastrophic learning. We also provide a theoretical analysis to guarantee the reliability and effectiveness of resistance to catastrophic forgetting.
  \item We validate ATS through real-world experiments on the RM65B robot, demonstrating its strong resistance to catastrophic forgetting and confirming its practical feasibility and advantages in real-world deployment scenarios.
\end{itemize}

\section{Related Work}
\subsection{Embodied Foundation Models}
Inspired by the training paradigm of “pre-training and fine-tuning” that has proven effective in LLMs, the training paradigm of robot foundation models has gradually evolved in this direction. With the release of a series of important research results, such as GR-2 from ByteDance Research\cite{1.1}, $\pi0$ launched by Physical Intelligence in 2024\cite{1.2}, the robot foundation model has formally entered into the era of “Pre-training to Fine-tuning”. RDT-1B from Tsinghua University, a diffusion foundation model for bimanual manipulation developed on the DiT framework\cite{2.4}. It standardizes the different action spaces of various robots by introducing a unified action representation, using a heterogeneous robot dataset containing more than 6000 trajectories for pre-training. In the fine-tuning phase, RDT-1B also encodes inputs from different modalities into a unified potential space to capture multi-modality in the action distribution. The model uses SigLIP as a visual encoder, T5-XXL as a verbal encoder, and diffusion Transformer to refine noisy action sequences into precise control actions based on the fused visual and text features\cite{1.3}. However, fine-tuning a model for a single task remains insufficient for generalization, and realizing continual learning across multiple tasks poses an ongoing challenge in robotics.

\subsection{Traditional Continual Learning}
Traditional continual learning aims to retain previously learned knowledge while updating the model with new task data. Existing traditional continual learning algorithms can be broadly classified into three categories\cite{2.5,2.6,2.7}. Regularization-based methods introduce extra constraints on parameters or regularize the model with prior data, such as limiting the learning rate on key parameters to mitigate the catastrophic forgetting. However, the memory budget will increase linearly as more and more tasks are learned, and the importance matrix of the parameters is difficult to determine and may conflict at different incremental stages. These factors affect the performance of regularization-based methods for training complex tasks\cite{2.8,2.9,2.10,2.11}. Architecture-based methods set up corresponding components for each task, which can be identified by expanding the network\cite{2.12,2.13} or by focusing on task-specific sub-networks\cite{2.14}. These methods rely on task identities to adjust how the networks operate, so there are significant limitations. Replay-based methods using memory playback techniques to retrain the model at the current step using past samples saved in the constructed data buffer or pseudo-samples generated by the generative model. Despite the simplicity of the concept, such methods have shown superior performance\cite{2.15,2.16,2.17}, and many recent studies have optimized them by exploiting self-supervised learning techniques\cite{2.18} or introducing knowledge distillation\cite{2.19}. However, these methods rely on the size of the buffer, which incurs a significant computational cost and is not applicable to tasks where data privacy should be considered.

\subsection{Continual Learning of Embodied AI}
Fine-tuning embodied foundation models pre-trained on large-scale data for a single task reduces computational cost. However, for high-generalization Embodied AI scenarios, robots need continuous learning of new skills, which cannot be achieved with minimal fine-tuning. Merging old and new datasets for retraining remains computationally expensive and data-intensive. Replay-based methods store a small amount of prior data, but task imbalance affects model performance as the number of tasks increases. LEGION, a replay-based method in robotics, is an incremental offline policy reinforcement learning framework for language embedding generation based on Bayesian nonparametric models. Its performance is validated only in structured scenarios with predefined tasks, and it lacks general knowledge for complex scenarios\cite{2.20}. In the field of LLMs, parameter-efficient fine-tuning\cite{2.21} (e.g., LoRA\cite{2.22}, adapter, prompt tuning) is effective, but its applicability to VLA models is unclear due to multi-modal data imbalance and challenges in determining parameter importance. Regularization-based continuous learning methods are also not applicable. Architecture-based methods rely on task identities, which are not obvious in robotic long-horizon operations involving multiple task types.

\section{The Proposed Method}
\subsection{Problem Definition}
Continual learning for embodied foundation models aims to enable agents to learn new task in dynamically changing environments while preserving the previously learned knowledge. Unlike traditional one-time training paradigms, this process is incremental and ongoing, allowing the model to evolve over time and ultimately become a general-purpose embodied agent capable of performing diverse tasks across a wide range of scenarios.

Denote a series of tasks as $\left\{D_1,\ldots,D_K\right\}$.For each task $D_k=\left\{\bm{X},\bm{Y}\right\}$, $\bm{X}=(\bm{X_P},\bm{X_I}, \bm{X_L})$, $\bm{Y}=(\bm{Y_l},\bm{Y_a})$, where $\bm{X_P} \bm{\in} \bm{R}^{N_k \times T \times d_P}$, $\bm{X_I} \bm{\in} \bm{R}^{N_k \times T \times c \times h \times w}$, $\bm{X_L} \bm{\in}\bm{R}^{N_k \times d_L}$, $\bm{Y_l} \bm{\in} \bm{R}^{N_k \times d_K}$, $\bm{Y_a} \bm{\in} \bm{R}^{N_k \times T \times d_a}$ denote the proprioception input, image input, language input, category label, and low-level actions for the $k-th$ task respectively; $N_k$ denotes the total amount of data for the  $k-th$ task; $T$ denotes the temporal length of the data; $d_P$, $d_a$ denote the dimension of proprioception and low-level actions;  $c,h,w$ denote the channels, height and width of the image input; $d_L$ denotes the length of language input; $d_K$ denotes the total number of categories. Given a model $\bm{M_\Theta}$ 
parameterized by $\bm{\Theta}=\left(\bm{\theta_1},\bm{\theta_2}\right)$, the goal of continual learning is to find an optimal set of parameters $\bm{\Theta}=\left(\bm{\theta_1},\bm{\theta_2}\right)$ that maximize the joint likelihood across all task datasets: 
\begin{equation}
    \max_{\bm{\Theta}}\sum_{k=1}^K\sum_{\bm{X},\bm{Y}\in D_k}[\log p_{\bm{\theta}_1}(\bm{Y_l}|\bm{X_L})\cdot\log p_{\bm{\theta}_2}(\bm{\epsilon}|\bm{X})]
\end{equation}
where $\bm{\epsilon}=\epsilon(\bm{Y_a})$ denotes the noise added to $\bm{Y_a}$.

Let $C_L$ denote the language input channel of the embodied foundation models, $M_{\theta_1}$ denotes the parameters of the analytic scheduler, and $g_k$ denotes the normalized weights of the analytic scheduler,  $k^*$ denotes the selected task identity, $C_{PIL}$ denotes the proprioception, vision and language channels,  $M_{\theta_2}$ denotes the parameters of the backbone of the embodied foundation model, $M_{\theta_2}^{k^*}$ denotes the parameters of the fine-tuned model specialized for task $k^*$, $\bm{Y_{a_{t+1}}}$ denotes the predicted low-level actions at time  $t+1$. At time step $t$, given the input $\bm{X_t}=(\bm{X_{P_t}},\bm{X_{I_t}}, \bm{X_L})$, the feedforward process can be expressed as follows: 
\begin{equation}
    k^*=\underset{k}{\operatorname*{\operatorname*{\arg\max}}} g_k(M_{\theta_1}(C_L(\bm{X_L})))
\end{equation}
\begin{equation}
    \bm{Y_{a_{t+1}}}={M_{\theta_2}}^{k^*}(C_{PIL}(\bm{X_t}))
\end{equation}

\subsection{Overall Architecture of ATS}
ATS introduces a shared-weight analytic scheduler and constructs a task-specific model library through single-task fine-tuning, enabling continual learning without replay while preserving the previously learned knowledge. This approach effectively mitigates catastrophic forgetting. The overall architecture of ATS consists of two modules: (a) Task-specific Model Library: This module comprises a set of independently fine-tuned models, each tailored to a specific task. This strategy not only effectively guarantees the performance of a single task, but also effectively avoids knowledge interference in multi-task training. With the continuous increase of tasks, the model library can be incrementally expanded, enabling ATS to continuously learn new tasks in open and dynamic environments. (b) Analytic Scheduler: This module introduces RLS in the language input channel to train an analytic scheduler, enabling continuously learning the mapping relationship between language instructions and task-specific models. Decoupled from the embodied foundation model, the analytic scheduler benefits from RLS’s inherent resistance to catastrophic forgetting, allowing it to continuously expand and adapt to new task. The overall architecture of ATS is illustrated in Figure 1. 

\begin{figure*}[htbp]
    \centering
    \begin{minipage}[b]{0.495\linewidth}\centering
        \includegraphics[width=\linewidth]{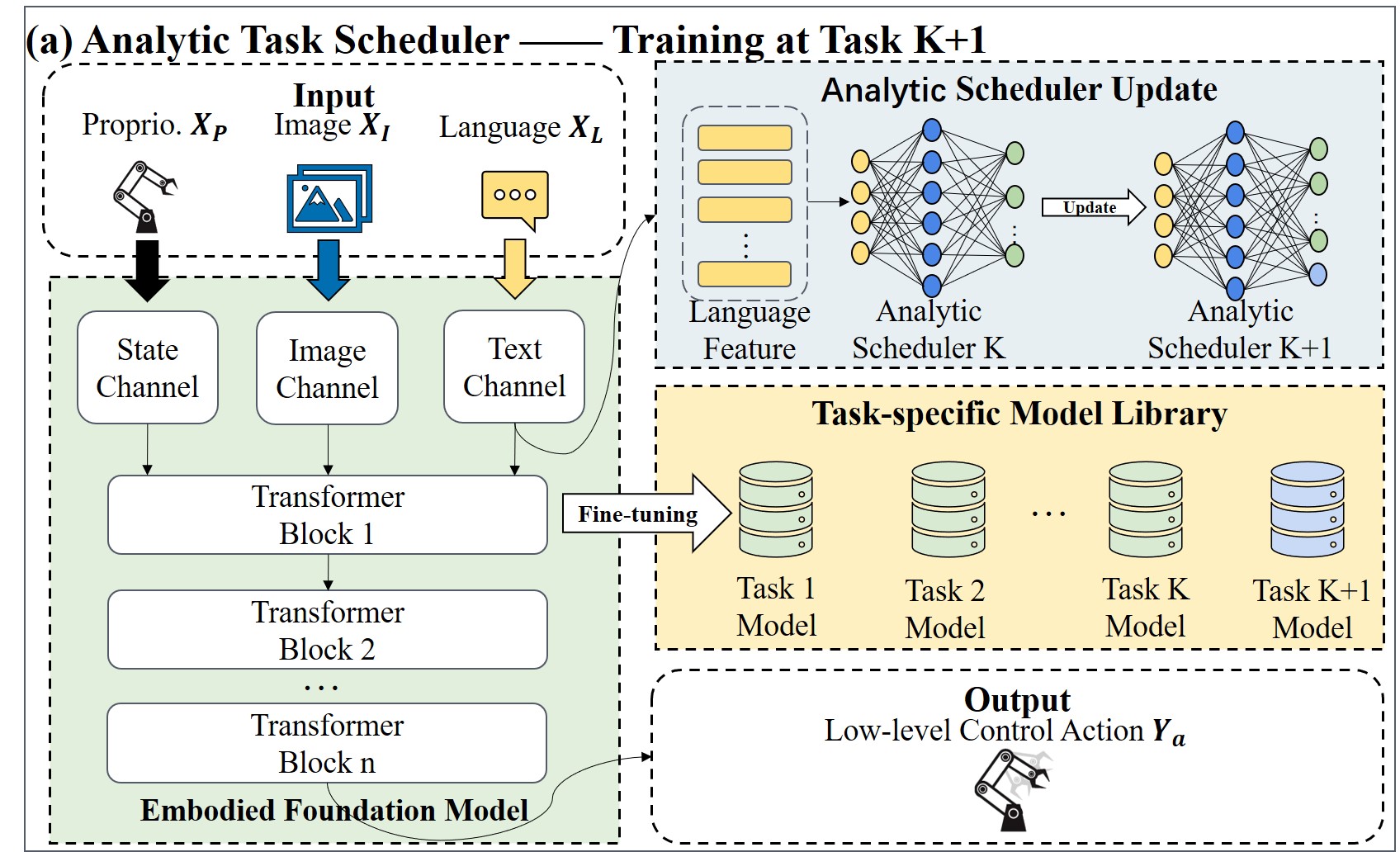}
    \end{minipage}
    \begin{minipage}[b]{0.495\linewidth}\centering
        \includegraphics[width=\linewidth]{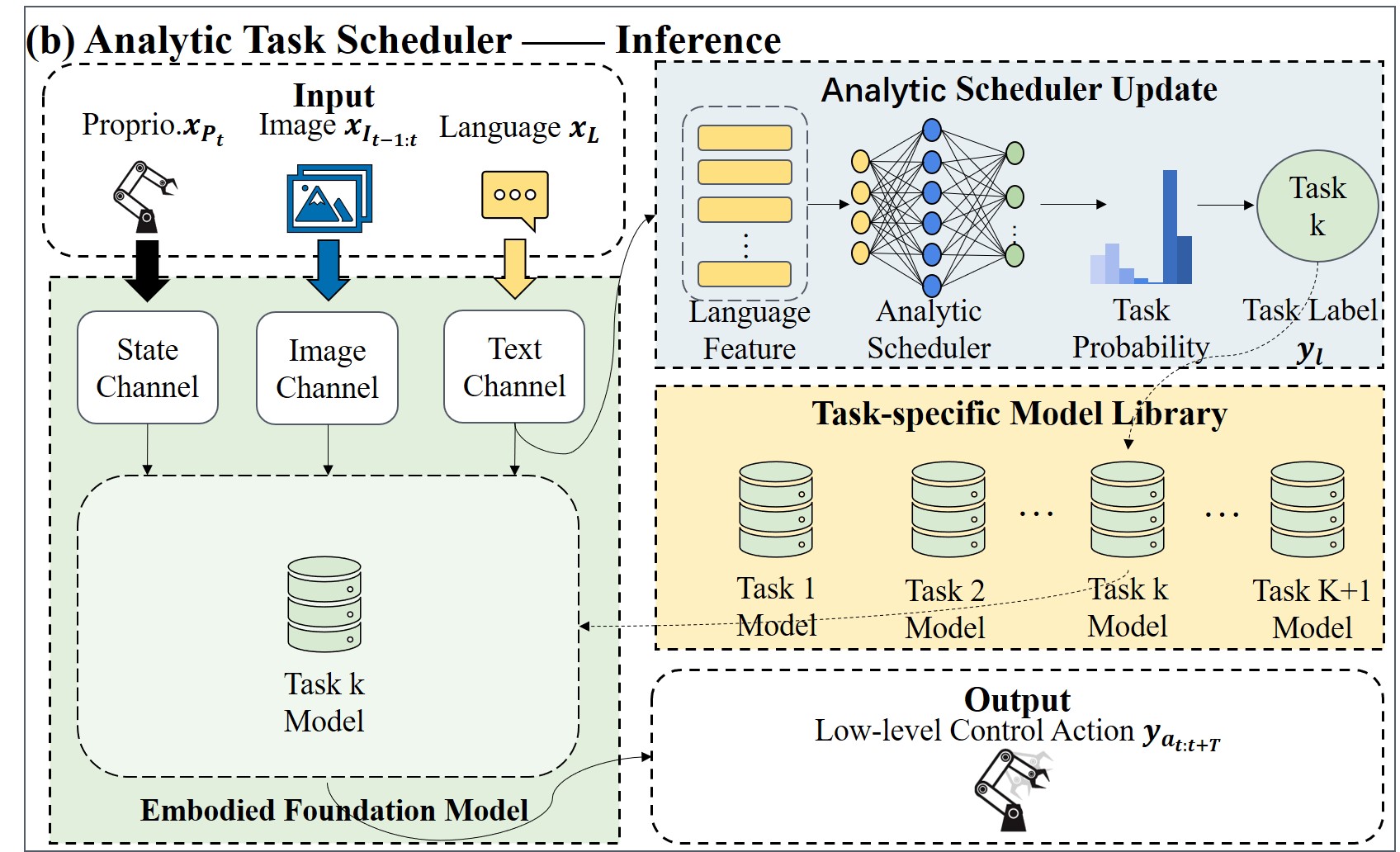}
    \end{minipage}
    \caption{An overview of ATS. (a) The training process of ATS involves two components: constructing task-specific model library and training an analytic scheduler. Task-specific model library comprises a set of fine-tuned models, each tailored to a specific task. The analytic scheduler introduces RLS in the language input channel, enabling continuously learning the mapping relationship between language instructions and task-specific models. 
    The language instructions will be extracted by $C_L$ and the extracted feature will be expanded to a larger dimension. Then the analytic scheduler will be trained in the way of ridge regression. The trained scheduler will perform recursive updates based on autocorrelation matrix $R_K$ and cross-correlation matrix $Q_K$, enabling efficient and forgetting-resistant learning. (b) The inference process of ATS. Given a language instruction, the analytic scheduler encodes the input and identifies the corresponding task by applying the learned weights. It then selects the appropriate model from the task-specific model library, which takes the multi-modal input and outputs the corresponding low-level actions to control the robot.}
    \label{fig1}
\end{figure*}

\subsection{Model Decoupling}

Since language is not only the most direct and natural medium for human-robot interaction, but also a rich source of semantic information and high-level task guidance that mitigates ambiguity in task intent, we decouple the language input channel of the embodied foundation model:
\begin{equation}
    C_L=W_{Text}
\end{equation}
\begin{equation}
    C_{PIL}=W_{State}+W_{Image}+W_{Text} 
\end{equation}
where $W_{Text}$,$W_{State}$ and $W_{Image}$ denote the weight parameters for the language, proprioception and image input channels, respectively.

\subsection{Task-specific Model Library}

For each task $D_k$, ATS adopts a single-task fine-tuning strategy to train a dedicated model $M_{\theta_2}^{k^*}$. Thus, for a sequence of tasks 
$\left\{D_1,\ldots,D_K\right\}$, we construct a task-specific model library $\left\{{M_{\theta_2}}^1,\ldots,{M_{\theta_2}}^{K}\right\}$. Unlike conventional continual learning approaches that incrementally train a single model across all tasks, our method independently fine-tunes a model for each task. This design effectively avoids parameter interference across tasks. Moreover, because each model is specifically optimized for a single task, the system can maintain high task performance while preserving previously learned knowledge. This independent fine-tuning strategy also offers greater flexibility and scalability for task transfer and adaptation across heterogeneous robotic platforms.

\subsection{Architecture of the Analytic Scheduler}
The analytic scheduler is designed to fulfill two objectives: (a) predict the task distribution $p(k|\bm{X_L})$ based on the language features $C_L(\bm{X_L})$  extracted from the language input channel of the embodied foundation model; and (b) enable continual learning of new tasks by updating weights in the absence of data from previous tasks, while effectively addressing the issue of catastrophic forgetting. To this end, we propose a ridge regression-based forgetting-resistant analytic scheduler strategy, which consists of two components: (a) a feature expansion module to enrich the representation capacity of language features, and (b) a RLS algorithm to support efficient and forgetting-resistant continual learning.

\textbf{\textbf{Feature Expansion Module.} }Given the language input, we first extract the sequential language features as: 
\begin{equation}
    L_{f_k}=C_L(\bm{X_L})\in R^{N_k\times N_L\times d_f}
\end{equation}
where $L_{f_k}$ denotes the extracted language features, $N_L$ denotes the sequence length, and $d_f$ is the dimension of each feature vector. 

To address task-specific variations in sequence length and simplify subsequent regression modeling, we apply average pooling along the temporal dimension to obtain a fixed-length feature representation:
\begin{equation}
    \bar{L}_{f_k}=mean(L_{f_k})=\frac1{N_L}\sum_{n=1}^{N_L}L_{f_n}\in R^{N_k\times d_f}
\end{equation}

To further enhance the expressive capacity of language features, we introduce a non-linear transformation by passing the pooled features through a fully connected layer with ReLU activation:
\begin{equation}
\tilde{L}_{f_k}=f_e(\overline{L}_{f_k})\in R^{N_k\times d_e}
\end{equation}
Here,  $f_e$ denotes a feedforward layer that maps the original features from dimension $d_f$ to a higher-dimensional space $d_e$ (where $d_f\ll d_e$),  thereby increasing the linear separability of task-specific features in the transformed space.

\textbf{Recursive Least Squares.} After feature expansion, ATS employs RLS to compute the weights of the analytic scheduler. Specifically, the expanded feature representations $\tilde{L}_{f_k}$ is linearly regressed to the corresponding task label matrix $Y_{l_{1:K}}$ by minimizing the following objective:
\begin{equation}
    \underset{W_{1:K}}{\operatorname*{\operatorname*{\arg\max}}}\left\|Y_{l_{1:K}}-\tilde{L}_{f_{1:K}}\cdot W_{1:K}\right\|_F^2+\gamma\|W_{1:K}\|_F^2
\end{equation}
where $\|\cdot\|_F$ denotes the Frobenius norm, $\gamma > 0$ denotes the regularization coefficient, $W_{1:K}\in R^{d_e \times {d_K}}$ denotes the analytic scheduler weights for tasks$\left\{D_1,\ldots,D_K\right\}$, $\tilde{L}_{f_{1:K}}\in R^{(\sum_{k=1}^K N_k)\times d_e}$ denotes the expanded feature matrix, and $Y_{l_{1:K}}\in R^{(\sum_{k=1}^K N_k)\times {d_K}}$ denotes the corresponding task labels.
\begin{equation}\tilde{L}_{f_{1:K}}=
\begin{bmatrix}
    \tilde{L}_{f_1} \\
    \tilde{L}_{f_2} \\
    \vdots \\
    \tilde{L}_{f_K}
    \end{bmatrix}\; Y_{l_{1:K}}=
    \begin{bmatrix}
    Y_{l_1} \\
    Y_{l_2} \\
    \vdots \\
    Y_{l_K}
\end{bmatrix}
\end{equation}

This convex problem admits a closed-form solution:
\begin{equation}
    \widehat{W}_{1:K}=\left(\sum_{i=1}^K \tilde{L}_{f_i}^T \tilde{L}_{f_i}+\gamma I\right)^{-1}\left(\sum_{i=1}^K\tilde{L}_{f_i}^T Y_{l_i}\right)
\end{equation}
where $I$ is a unit matrix.

Let us define the following auxiliary matrices:
\begin{equation}
R_K=\left(\sum_{i=1}^K\tilde{L}_{f_i}^T\tilde{L}_{f_i}+\gamma I\right)^{-1}\in R^{d_e\times d_e}
\end{equation}
\begin{equation}
    Q_K=\left(\sum_{i=1}^K\tilde{L}_{f_i}^T Y_{l_i}\right)R^{d_e\times {d_K}}
\end{equation}

Substituting Equation (12) and Equation (13) into the closed-form solution Equation (11), we have:
\begin{equation}
    \widehat{W}_{1:K}=R_KQ_K\in R^{d_e\times {d_K}}
\end{equation}
where $R_K$ is the autocorrelation matrix, storing intrinsic relationships within the feature matrix $\tilde{L}_{f_{1:K}}$, while $Q_K$ is the cross-correlation matrix that stores the mapping between language feature matrix $\tilde{L}_{f_{1:K}}$ and task labels matrix $Y_{l_{1:K}}$. During incremental learning of a new task $\left\{D_{K+1}\right\}$, the analytic scheduler can update its weights $\widehat{W}_{1:K}$ using only $R_K$ and $Q_K$ without revisiting the data from previous tasks. This enables efficient and replay-free continual learning.

\subsection{Incremental Update of the Analytic Scheduler}

According to the Woodbury matrix identity\cite{3.1}, for any invertible matrices $A$ and $C$, the following holds:
\begin{equation}
    (A+UCV)^{-1}=A^{-1}-A^{-1}U(C^{-1}+VA^{-1}U)^{-1}VA^{-1}
\end{equation}

Let $A={R_K}^ {-1}$, $U =\tilde{L}_{f_{1:K}}^T$, $V=\tilde{L}_{f_{1:K}}$, and $C=I$ and substituting into Equation (15), we obtain: 
\begin{equation}
\begin{aligned}
    &R_{K+1} = \left({R_K}^{-1} + {\tilde{L}_{f_{K+1}}}^T \tilde{L}_{f_{K+1}}\right)^{-1} \\
    &= R_K - R_K \tilde{L}_{f_{K+1}}^T \left(I + \tilde{L}_{f_{K+1}} R_K \tilde{L}_{f_{K+1}}^T\right)^{-1} \tilde{L}_{f_{K+1}} R_K
\end{aligned}
\end{equation}

From Equation (14), the solution to the RLS-based optimization is:
\begin{equation}
\widehat{W}_{K+1}=R_{K+1}Q_{K+1}=R_{K+1}Q_K+R_{K+1}\tilde{L}_{f_{K+1}}^T Y_{l_{K+1}}
\end{equation}

Substituting Equation (16) into Equation (17), we obtain the closed-form expression:
\begin{equation}
\begin{aligned}
    \widehat{W}_{K+1} & =\left(I-R_{K+1}\tilde{L}_{f_{K+1}}{\tilde{L}_{f_{K+1}}}^T \right)\widehat{W}_{1:K} \\
     & +R_{K+1}\tilde{L}_{f_{K+1}}^TY_{l_{K+1}}
\end{aligned}
\end{equation}

Assuming a new task $\left\{D_{K+1}\right\}$, and given the previously computed matrices $R_K$ and $C_K$, the updated analytic scheduler weights $\widehat{W}_{K+1}$ for the new task can be directly calculated using Equation (18). Since the update is performed through analytical computation, it does not require access to historical task data, thereby achieving complete resistance to catastrophic forgetting.

\subsection{Inference Procedure and Task Execution Pipeline}

At time step $t$, given the input $\bm{x_t}= (\bm{x_{P_t}},\bm{x_{I_t}},\bm{x_L})$,  the language feature $l_f$ is extracted through the language input channel:
\begin{equation}
    l_f=C_L(\bm{x_{L}})
\end{equation}

$l_f$ is then passed through the feature expansion module to obtain the enhanced representation $\tilde{l}_{f}$ :
\begin{equation}
    \tilde{l}_f=f_e(\mathrm{mean}(l_f))
\end{equation}

Using linear regression on $\tilde{l}_{f}$ , the probability distribution over task categories is computed as:
\begin{equation}
    p(k|\bm{x_L})=\mathrm{softmax}(\tilde{l}_f\widehat{W}_{K+1})
\end{equation}

The task category with the highest predicted probability is selected:
\begin{equation}
    k^*=\underset{k}{\operatorname*{\operatorname*{\arg\max}}}p(k|\bm{x_L})
\end{equation}

Based on the identified task $k^*$, the corresponding model is retrieved from the task-specific model library and executed:
\begin{equation}
    \bm{y_{a_{t:t+T}}}={M_{\theta_2}}^{k^*}(\bm{x_t})
\end{equation}
\begin{table}[h]
  \centering
  \begin{tabular}{ll}
    \toprule
    \textbf{Algorithm 1} The training process of the Analytic Scheduler \\
    \midrule
    \textbf{Analytic Scheduler training for} $D_1, \ldots, D_K$. \\
    1. Extract features $C_L(\bm{X_L})$ via (6) \\
    2. Expand features to $\tilde{L}_{f_{1:K}}$ via (7) and (8). \\
    3. \textbf{for} $k=1$ \textbf{to} $K$ \textbf{do}: \\
       \hspace{2em} Fine-tuning model $M_{\theta_2}{}^k$. \\
       \hspace{1em}\textbf{end for} \\
    4. Obtain analytic router weight $\widehat{W}_{1:K}$ via(11). \\
    5. Save the AutoCor $R_K$ and CrossCor $Q_K$. \\
    \textbf{Continual Learning for} $D_{K+1}$. \\
    1. Extract features $C_L(\bm{X_L})$ via (6) \\
    2. Expand features to $\tilde{L}_{f_{K+1}}$ via (7) and (8). \\
    3. Fine-tuning model $M_{\theta_2}{}^{K+1}$. \\
    4. Obtain analytic scheduler weight $\widehat{W}_{K+1}$ via(18). \\
    5. Save the AutoCor $R_{K+1}$ and CrossCor $Q_{K+1}$. \\
    \bottomrule
  \end{tabular}
\end{table}
\begin{table}[h]
  \centering
  \begin{tabular}{ll}
    \toprule
    \textbf{Algorithm 2} The inference process of the Analytic Router \\
    \midrule
    \textbf{Inference} $\bm{x_t}= (\bm{x_{P_t}},\bm{x_{I_t}},\bm{x_L})$. \\
    1. Extract features $C_L(\bm{x_L})$ via (19). \\
    2. Expand features to $\tilde{l}_f$ via (20). \\
    3. Calculate task probability via (21). \\
    4. Select task-specific model via (22). \\
    5. Forward propagation via (23). \\
    \bottomrule
  \end{tabular}
\end{table}

\section{Experiments}

To validate the effectiveness of the proposed ATS in embodied tasks, we construct a multitask evaluation environment designed to assess both the continual learning capability of the analytic scheduler and the task adaptability of individually fine-tuned models.
\subsection{Benchmarks}

Our experimental setup includes a language dataset for training the analytic scheduler and a robot manipulation dataset for fine-tuning the embodied foundation model. 

\textbf{Hardware Information:}The experiments are conducted on a RealMan RM65B robotic arm, equipped with an Intel RealSense D435 wrist camera and an exterior Intel RealSense D435i camera, as shown in Figure 2.

\textbf{Language Dataset:} We design the dataset comprising 10 categories of language instructions, each reflecting a typical embodied scenario. These include tasks such as picking up a banana, stacking tomatoes, placing a stapler, selecting a die, pouring half a glass of water, picking up a cube, redirecting a Rubik’s cube, controlling a robotic dog, stacking cans, and picking up corn. Each category contains 102 unique textual instructions that are manually crafted and validated to ensure language diversity and clarity of intent. The dataset is employed to train the analytic scheduler and evaluate the performance of task recognition.

\textbf{Robot Manipulation Dataset: }We focus on two manipulation scenarios: picking up the banana and picking up the corn, as shown in Figure 3 Figure 4. For each task, 20 complete execution episodes are collected. Each episode includes proprioception, images from an external camera and a wrist camera, the corresponding language instructions, and a sequence of low-level actions. All episodes were collected using a real-world robot, RM65B, capturing the complexity and variability inherent in embodied task execution.
\begin{figure}[!htb]
  \centering
  \includegraphics[width=\hsize]{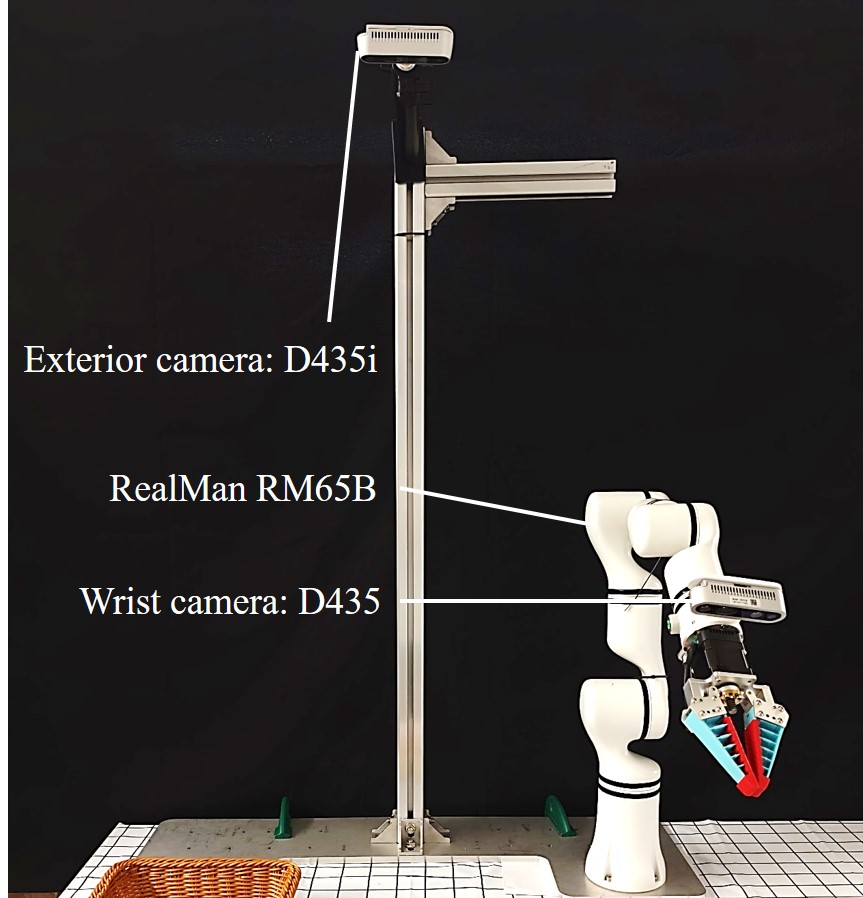}
  \caption{Hardware features.}
  \label{fig2}
\end{figure}

\begin{figure}[!htb]
  \centering
  \includegraphics[width=\hsize]{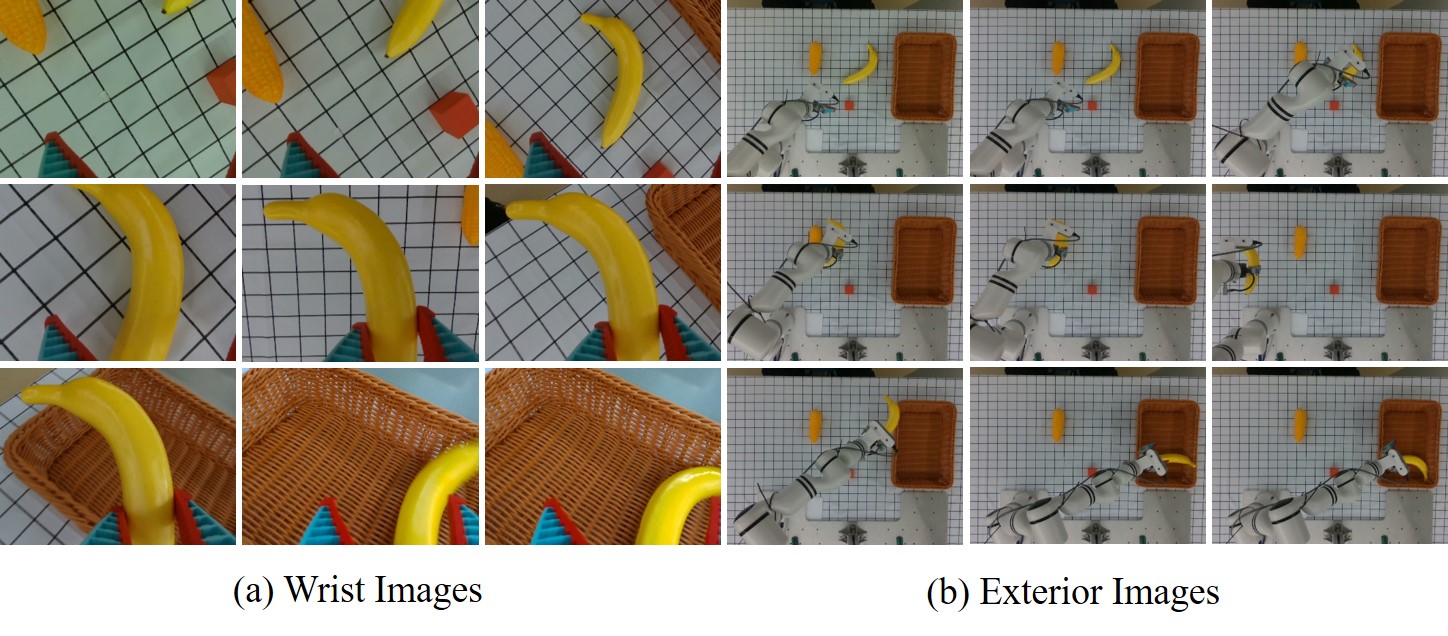}
  \caption{Pick up the banana and place it into the basket.}
  \label{fig3}
  \includegraphics[width=\columnwidth]{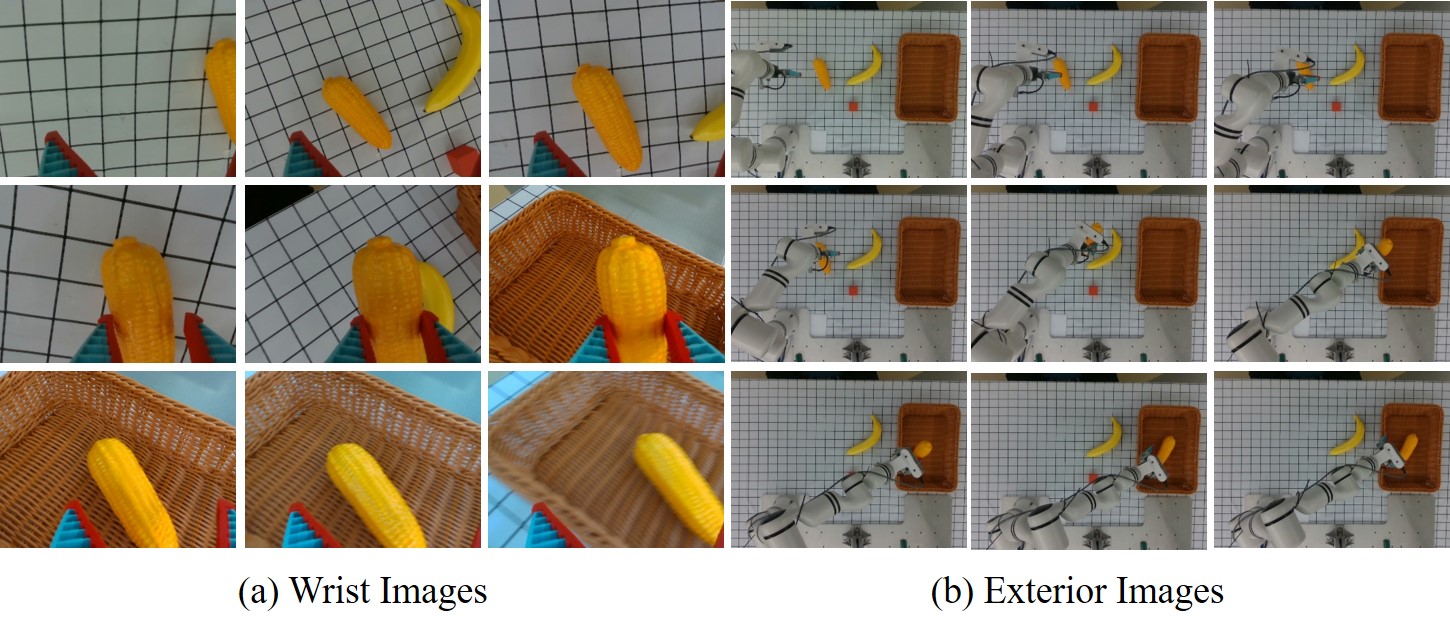}
  \caption{Pick up the corn and place it into the basket.}
  \label{fig4}
\end{figure}

\subsection{Evaluation Metrics}

To comprehensively evaluate the performance of our proposed ATS, we adopt three evaluation metrics: average accuracy, forgetting rate, and task execution score.

\textbf{Average Accuracy: }Average Accuracy measures the overall performance of the analytical scheduler in continual learning settings. It is defined as:
\begin{equation}
    \bm{\bar{A}}=\bm{\frac{1}{K+1}\sum_{k=1}^{K+1}A_{k}}
\end{equation}
where $\bm{A_k}$ denotes the accuracy of the scheduler on the cumulative test set ${D_{0:k}}^{test}$ after the $k-th$ learning phase. A higher value of $\bar{A}$ indicates better generalization and stronger retention of previous tasks while learning new ones.

\textbf{Forgetting Rate: }Forgetting Rate quantifies how much the model forgets previous tasks after learning new ones. It is computed as:
\begin{equation}
    \bm{F_k=\tilde{A}_1-\tilde{A}_k}
\end{equation}
where $\bm{\tilde{A}_1}$ is the initial accuracy on task \textbf{1} and $\bm{\tilde{A}_k}$ is the accuracy on the same task after the learning phase $k$. Lower values of $F_k$ indicates a stronger resistance to catastrophic forgetting.

\textbf{Task Execution Score: }To assess the embodied performance of the fine-tuned models, we design a structured scoring protocol for each task, with a maximum score of 100 points. The score is evenly distributed across four sub-goals, as shown in Table 1. An example of picking up the banana is shown in Figure 5.

\begin{table}[h]
    \centering
    \caption{Task Execution Scoring Criteria }
    \begin{minipage}{\textwidth}
        \begin{tabular}{ccc}
            \hline
            & Description & Score \\
            \hline
            Stage 1 & Move correctly to the target object. & 20 \\
            Stage 2 & Reach above the target object. & 30 \\
            Stage 3 & Pick up the target object. & 30 \\
            Stage 4 & Place the object at the designated location. & 20 \\
            \hline
        \end{tabular}
    \end{minipage}
\end{table}

\begin{figure}[!htb]
  \centering
  \includegraphics[width=\columnwidth]{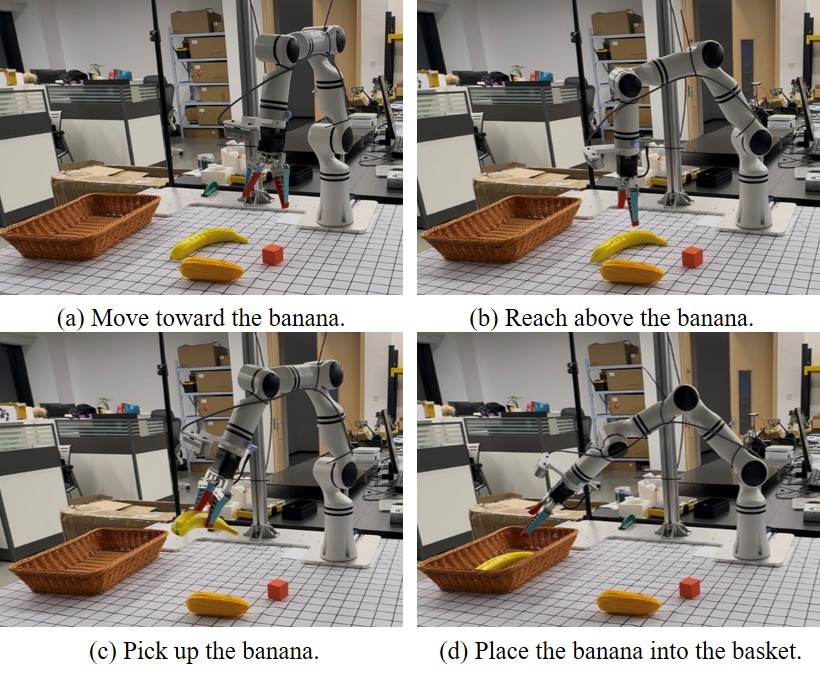}
  \caption{Task Execution Sub-goals}
  \label{fig5}
\end{figure}

If the robot mistakenly picks up the wrong object, a penalty of 25 points is subtracted from the total score. Each model is tested across 20 trials, with a maximum inference time of 2 minutes per trial. The final Task Execution Score is computed as the average score over these trials, reflecting the model's real-world task competence.

\subsection{Training details}
Our experiments are conducted with the RDT-170m embodied foundation model as the backbone. Both the analytic scheduler and task-specific model library are implemented in PyTorch and trained on two NVIDIA RTX 4090 GPUs, each equipped with 24GB of VRAM.

\textbf{Analytic Scheduler: }The scheduler is trained to perform task recognition based on language inputs. A two-phase incremental learning protocol is designed for training the scheduler to recognize tasks from language inputs, involving 10 task categories in total. In the base training phase, the model is trained on 5 initial task categories. The remaining 5 categories are sequentially introduced, one per phase, across five incremental learning (IL) phases. A batch size of 256 is used throughout all training stages.

As shown in Table 2, the scheduler achieves near-perfect performance across all stages of continual learning. As shown in the table below, the model maintains 98.61\% accuracy up to IL Phase 4 and experiences only a slight drop (to 95.83\%) in the final phase. Correspondingly, the average forgetting rate remains close to 2.34\%, indicating excellent retention of previously learned tasks.
\begin{table}[h]
    \centering
    \caption{Continual Learning Accuracy and Forgetting Rate}
    \begin{tabular}{@{}ccc@{}} 
        \hline
        & \multicolumn{1}{c}{Accuracy (\%)} & \multicolumn{1}{c}{Forgetting Rate (\%)} \\
        \hline
        Base training & 100 & 0.00 \\
        IL Phase 1 & 98.61 & 1.67 \\
        IL Phase 2 & 98.61 & 1.67 \\
        IL Phase 3 & 98.61 & 1.67 \\
        IL Phase 4 & 98.61 & 1.67 \\
        IL Phase 5 & 95.83 & 5.00 \\
        \hline
    \end{tabular}
\end{table}

\textbf{Task-specific Model Library: }We fine-tune the RDT-170m model separately for two manipulation tasks: “Pick up the banana” and “Pick up the corn”. Fine-tuning is conducted in single-task way using a batch size of 16, a learning rate of 1e-4, and a maximum of 200,000 steps. All training is carried out on the same GPU configuration as above.

To examine the effects of task interference and forgetting, we evaluate several configurations:
\begin{itemize}
  \item \textbf{B after A}: Task B is fine-tuned after Task A.
  \item \textbf{mix}: Both tasks are fine-tuned simultaneously in a multi-task setting.
  \item \textbf{ATS}: Task-specific models are selected and executed based on the output of the analytic scheduler.
\end{itemize}

The following results illustrate the impact of different fine-tuning strategies on task execution score:
\begin{table}[h]
    \centering
    \caption{Task Execution Scores}
    \begin{tabular}{@{}ccc@{}}
        \hline
        & Score \\
        \hline
        Pick up the banana (corn after banana) & 23.05 \\
        Pick up the corn (corn after banana) & 48.33  \\
        Pick up the banana (mix) & 50.71 \\
        Pick up the corn (mix) & 45.71 \\
        Pick up the banana (ATS) & 66.27 \\
        Pick up the corn (ATS) & 66.50 \\
        \hline
    \end{tabular}
\end{table}

As shown in Table 3, the traditional fine-tuning strategy suffers from significant forgetting that fine-tuning on a new task severely degrades the performance on the previously learned one. Joint fine-tuning introduces noticeable interference between tasks, resulting in suboptimal performance for both. In contrast, the ATS framework retains task-specific performance with minimal forgetting, enabling both tasks to be performed robustly.

\subsection{Experimental Result Analysis}

\textbf{Single-Task Fine-Tuning vs. Multi-Task Fine-Tuning:} To assess the impact of task interference on model performance, we conduct experiments using two tasks under both single-task and joint fine-tuning settings. In the single-task setting, each task is fine-tuned independently to construct dedicated task-specific model in the task model library. In contrast, the joint fine-tuning setting uses data from both tasks simultaneously to train a shared model.

Results show that single-task fine-tuning leads to stable and efficient task execution, with each model demonstrating strong specialization and robustness on its respective task. However, when multiple tasks are trained jointly in an attempt to improve generalization, performance on both tasks degrades significantly. This degradation highlights a strong interference effect: shared parameters optimized for one task tend to overwrite or disrupt the representations learned for another, making it difficult for the model to balance performance across tasks. The findings suggest that conventional multi-task fine-tuning is prone to parameter conflict and is ill-suited for continual learning settings in embodied AI.

\textbf{Catastrophic Forgetting in Sequential Fine-Tuning: }To further investigate the stability of traditional fine-tuning strategies in continual learning, we apply sequential fine-tuning, where a model trained on one task is subsequently fine-tuned on another, without access to the previous task’s data. While this method maintains acceptable performance on the new task, it results in nearly complete performance degradation on the original task, exhibiting severe catastrophic forgetting.

This outcome underscores a critical limitation of conventional fine-tuning: in the absence of replay data or structural protection, parameter updates for new tasks tend to overwrite knowledge learned from previous ones. This is particularly problematic in scenarios where data replay is impractical due to privacy, memory, or compute constraints. Such instability renders traditional fine-tuning strategies inadequate for real-world embodied continual learning applications.

\textbf{Effectiveness of the ATS Framework: }The introduction of the Analytic Task Scheduler (ATS) framework significantly mitigates both task interference and forgetting in continual learning. ATS decomposes the problem into two components: a task-specific model library, where each model is tailored to a specific task, and a lightweight analytic scheduler that selects the appropriate model based on the  language instruction input.

By avoiding parameter sharing across tasks, ATS eliminates direct cross-task interference, preserving each model’s task-specific capabilities. The analytic scheduler, trained on language instructions, accurately identifies the task category and routes execution to the correct task-specific model. This design ensures high decision accuracy and robust task scheduling even in continual learning scenarios.

Moreover, the backbone of analytic scheduler is built upon ridge regression and RLS algorithm, enabling resistance to catastrophic forgetting. It requires only statistics—namely, autocorrelation and cross-correlation matrices—to update weights, and does not rely on previous task data. This allows the scheduler to achieve forgetting-free updates under non-replay conditions, making ATS uniquely suited for scalable continual learning.

Finally, the compact architecture of analytic scheduler allows it to operate efficiently without GPU acceleration. Its inference speed ensures compatibility with edge devices and resource-constrained deployment environments. Overall, ATS offers a practical and effective solution for continual learning in embodied foundation models.

\section{Conclusion}

This paper presents the Analytic Task Scheduler (ATS), a continual learning framework designed to learn new task in a replay-free way without destroying the original generic knowledge of the model, while effectively overcoming catastrophic forgetting. ATS integrates a lightweight, recursive least squares-based scheduler into the language input channel to enable accurate task recognition and model selection. Combined with a task-specific model library constructed through single-task fine-tuning, ATS fundamentally avoids parameter conflicts across tasks and significantly enhances the model’s continual learning capabilities.

The proposed approach offers several key advantages:

\textbf{Forgetting-Resistant Learning: }By decoupling the model structure and leveraging recursive least squares for weight updates, ATS effectively prevents catastrophic forgetting during continual learning.

\textbf{Replay-Free Adaptation: }The scheduler relies solely on sufficient statistics—the autocorrelation and cross-correlation matrices—for incremental updates, enabling data-efficient learning without requiring access to historical task data.

\textbf{Interference-Free Fine-Tuning: }Each task is fine-tuned independently, resulting in a dedicated sub-model per task. This eliminates cross-task parameter interference and improves both individual task performance and system stability.

\textbf{Lightweight and Deployable: }The analytic scheduler has a minimal computational footprint, requires low memory, and supports fast inference even on CPU-only setups, making it well-suited for edge and resource-constrained environments.

In summary, ATS provides a general, efficient, and deployable solution for continual learning in embodied foundation models in complex and dynamic task environments. Its scalability, stability, and adaptability make it a promising direction for future research and real-world applications.

\end{document}